\documentclass{article}
\usepackage[utf8]{inputenc}

\usepackage{amsmath}
\usepackage{graphicx}
\graphicspath{ {./figures/} }
\usepackage{biblatex}
\title{Using the discrete radon transformation for grayscale image moments}
\author{William Diggin
     \and 
     Michael Diggin}
\date{August 2020}

\begin{document}

\maketitle

\begin{abstract}
Image moments are weighted sums over pixel values in a given image and are used in object detection and localization. Raw image moments are derived directly from the image and are fundamental in deriving moment invariants quantities.
The current general algorithm for raw image moments is computationally expensive and the number of multiplications needed scales with the number of pixels in the image. For an image of size $(N,M)$, it has $O(N \cdot M)$ multiplications. In this paper we outline an algorithm using the Discrete Radon Transformation for computing the raw image moments of a grayscale image. It reduces two dimensional moment calculations to linear combinations of one dimensional moment calculations. We show that the number of multiplications needed scales as $O(N+M)$, making it faster then the most widely used algorithm of raw image moments. 
\end{abstract}

\section{Introduction}
Image moments were first introduced by Hu \cite{HuArticle}, in the context of pattern recognition. Hu defined different moments, invariant under translations, rotations and scaling.  Image moments are useful for describing objects after segmentation and have been applied to various image analysis problems. 

In particular it has been applied to problems in healthcare. In 2015 it was applied to solve the Pathological Brain Detection problem \cite{zhang2015pathological}, even more recently in 2020 image moments were used to extract information about object orientation from micro-X-ray tomography image data \cite{doerr2020micro}.

Computing image moments is an expensive operation and generally scales with the number of pixels. It comprises of computing sums over the pixel values (typically integer values between 0 and 255), weighted by a power of the distance from the start or origin of the image. 
For small images this isn't an issue, but as images are getting larger and larger, with many now in 4k resolution, the computation time is scaling very poorly. The problem is that as image size increases linearly, the number of pixels and thus the computation time increases quadratically. The algorithm we propose here scales instead with roughly the square root of the number of pixels, or rather with the width and height of the image.
Of course, there are some select algorithms that compute image moments exceptionally fast, in certain cases, e.g. for a binary image \cite{tomavs1946calculation}. 

In section 2 we give an introduction to raw image moments, the base computations that are used to derive the moment invariants. In section 3 we introduce the Discrete Radon Transformation and derive how to compute the raw image moments using this algorithm, showing how it reduces the number of multiplications required dramatically. Finally, in section 4, we illustrate computation times of the proposed algorithm against the current algorithm that is it widely used and part of the OpenCV library \cite{opencv_library}.

\section{Raw Image Moments}

These invariant moments from Hu \cite{HuArticle} are built upon more fundamental quantities, called \textit{raw moments}, that are derived directly from an image or polygon shape. Given an array of pixels \begin{math} I \end{math} (a grayscale image), a raw moment of order $(i, j)$ is calculated as 
\begin{equation}
\label{base_moments}
    M_{ij} = \sum_{x, y} I(x,y) x^i y^j 
\end{equation}
In order to derive the seven Hu invariants \cite{HuArticle}, only the first ten raw moments are needed: $ M_{00}, M_{01}, M_{10}, M_{02}, M_{11}, M_{20}, M_{03}, M_{12}, M_{21} \text{ and } M_{30}$.  

As can be seen from \eqref{base_moments}, this calculation is expensive in terms of multiplications. If the pixel array $I$ has height and width $(N, M)$, then each moment has $O(N\cdot M)$ multiplications.

In particular, the method that we will be comparing against is the method currently used by the popular open source computer vision library OpenCV \cite{opencv_library}. For a grayscale image, this method performs $3N\cdot M + 7N$ multiplications in calculating all of the raw moments. It is more optimized than the naive implementation of \eqref{base_moments} but still scales with the total number of pixels in the image.

\section{The Discrete Radon Transformation}

Current methods for computing the raw moments of a grayscale pixel array involve calculating multiple 2 dimensional moments. The Discrete Radon Transformation (DRT) \cite{DRT} reduces the computation from 2D moments of one 2D array, to 1D moments of 4 1D arrays. The original raw moments are linear combinations of these 1D moments. This is what enables this method to have $O(N+M)$ multiplications. The radon transform (not in the discrete form) has been used before in terms of optical feature extraction, but from a hardware perspective \cite{gindi1984optical}. In the discrete form, it was looked at for image moment calculations before in \cite{shen1996fast}. They similarly found this method to be faster. However in this paper we give a derivation of the how the transformation gives the exact raw image moments, as well as compare it against the current best algorithm implemented for grayscale images, showing that it is still significantly faster for computations.  

The four one dimensional arrays mentioned above are projections of the image, vertically, horizontally, and at two angles, 45 degrees and 135 degrees (diagonally and anti-diagonally). The pixels are summed along these axes. Concretely, defining the four \textit{projection arrays} as $V, H, D, A$, we have:

\begin{align}\label{projection_def}
    V[k] &= \sum_{j} I(k, j)\\
    H[k] &= \sum_{j} I(j, k)\\             
    D[k] &= \sum_{i+j=k} I(i, j)\\         
    A[k] &= \sum_{j-i+M-1=k} I(i, j) 
\end{align}

\begin{figure}
\centering
\includegraphics[scale=0.5]{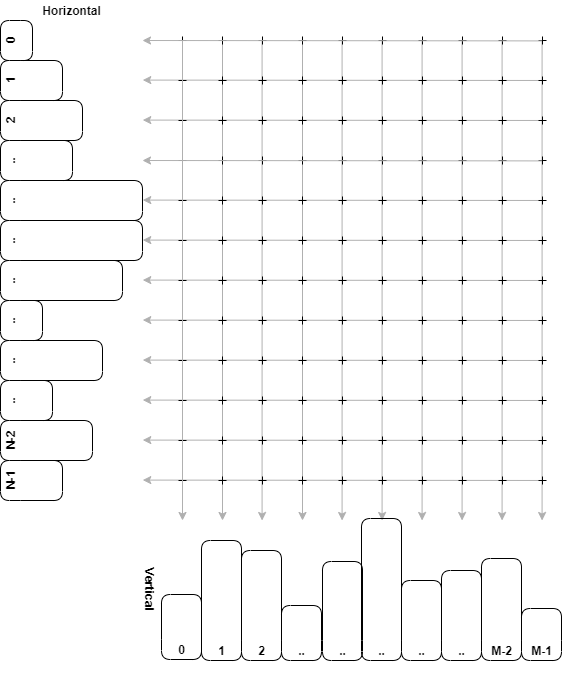}
\caption{Figure illustrating the vertical and horizontal projections}
\label{fig:plot1}
\end{figure}

\begin{figure}
\centering
\includegraphics[scale=0.25]{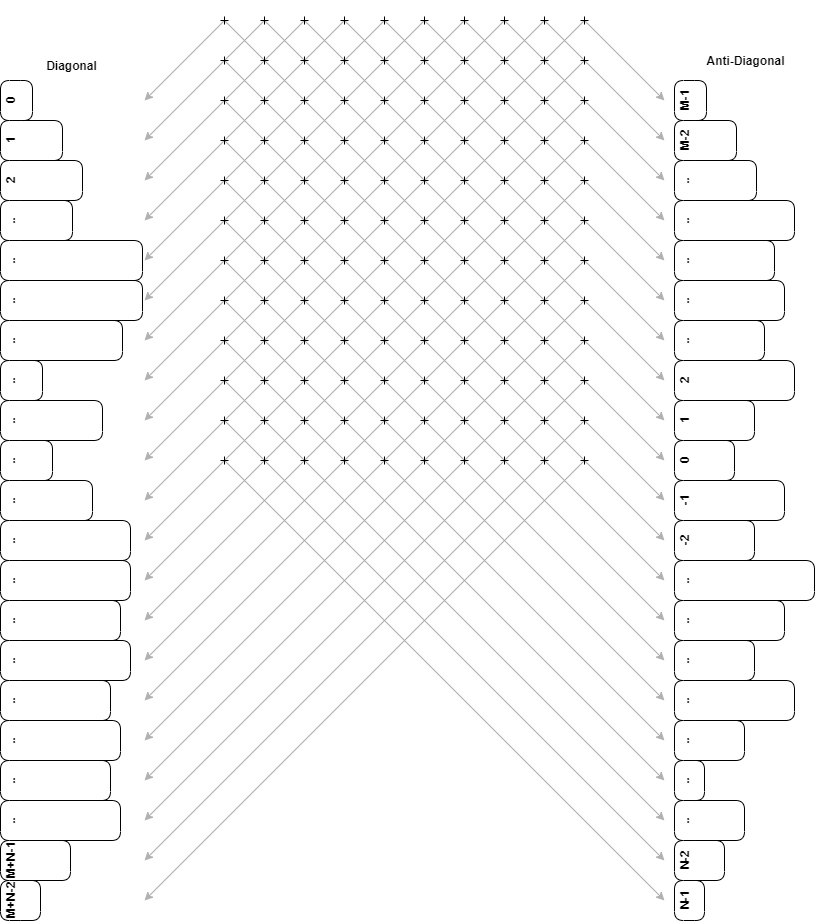}
\caption{Figure illustrating the diagonal and anti-diagonal projections}
\label{fig:plot2}
\end{figure}

Where $V[k]$ denotes the $k^{th}$ element of the vertical projection array, and similarly for the other three arrays. Illustrations of these projections are in figure \ref{fig:plot1} and figure \ref{fig:plot2}. The strange summation condition on $A[k]$ is to ensure that the array starts with index zero and has the right length.  $V, H, D, A$ have lengths $M, N, N+M \text{ and } N+M$ respectively for an image with height and width $N, M$. In the Appendix we show examples of the projections for an image.

The raw moments can then be constructed with the moments of these arrays. As an example, we denote $V_i$ as the $i^{th}$ moment of the array $V$.
\begin{equation}\label{arraymoment}
    V_i = \sum_k V[k] \cdot k^i
\end{equation}
The one exception to this is the anti-diagonal array. Due to it's definition (in particular, the way that it iterates over the pixels), we define 
\begin{equation}\label{antimoment}
    A_i = \sum_{k} A[k]\cdot (k-M+1)^{i}
\end{equation}

We claim that the true exact raw image moments from the original pixel array can be computed as
\begin{align}
    M_{00} &= V_0 \\
    M_{i0} &= V_i \\
    M_{0i} &= H_i \\
    M_{11} &= (D_2 - M_{20} - M_{02})/2 \\
    M_{12} &= (D_3 + A_3)/6 - M_{30}/3 \\
    M_{21} &= (D_3 - A_3)/6 - M_{03}/3
\end{align}

The proof of $M_{i0}$ for $i >= 0 $ follows from \eqref{arraymoment} and uses \eqref{base_moments}
\begin{align*}\label{proof1}
    V_i &= \sum_k V[k]k^i \\
    &= \sum_k (\sum_j I(k, j)) k^i \\
    &= \sum_{k, j} I(k, j) k^i \\
    &= M_{i0}
\end{align*}

The proof for $M_{0i}$ from $H_i$ is very similar.
For $M_{11}$ the argument goes as follows, we start with $D_2$
\begin{align*}
    D_2 &= \sum_{k=0}^{N+M-1} D[k] \cdot k^2 \\
    &= \sum_{k=0}^{N+M-1} \sum_{i+j=k} I(i, j) \cdot k^2 \\
    &= \sum_{i, j} I(i, j) \cdot (i+j)^2 \\
    &= \sum_{i, j} I(i, j) \cdot (i^2 + j^2 + 2ij) \\
    &= M_{20} + M_{02} + 2 M_{11}
\end{align*}
Rearranging gives the result in equation (11).

Similar to the above, it can be shown that $D_3 = M_{30} + M_{03} + 3M_{21} + 3M_{12}$. We also show that 
\begin{align*}
    A_3 &= \sum_{k} A[k] \cdot (k-m+1)^3 \\
    &= \sum_{k} \sum_{j-i=M-1+k} I(i, j) \cdot (k-M+1)^3 \\
    &= \sum_{i, j} I(i, j) \cdot(j-i)^3 \\
    &= M_{03} - M_{30} + 3M_{21} - 3M_{12}
\end{align*}
Combining this result for $A_3$ with the above result for $D_3$, they can be rearranged to produce equations (12) and (13), completing the proof. 

The four projection arrays have length of at most $N+M$, therefore each inner product has at most $N+M$ multiplications, leading to the O(N+M) complexity. In fact, the number of multiplications can be found exactly. The power arrays that are used in the moment calculations (arrays of $k^i$ for $0 \leq k \leq N+M-1$, for $i = [0, 1, 2, 3]$) can be computed ahead of time and stored. This leaves the algorithm with just $6(N+M)$ multiplications.

\section{Comparison and Results}

We compare the DRT method with the current method implemented as part of the OpenCV library \cite{opencv_library} on 2D grayscale images of increasing size. All computations were carried out on an Intel Core i5 7th Gen processor, and results were taken as the fastest of 1000 runs. We have also added in the results for the naive implementation \eqref{base_moments}.

\begin{figure}[ht]
\includegraphics[width=\textwidth]{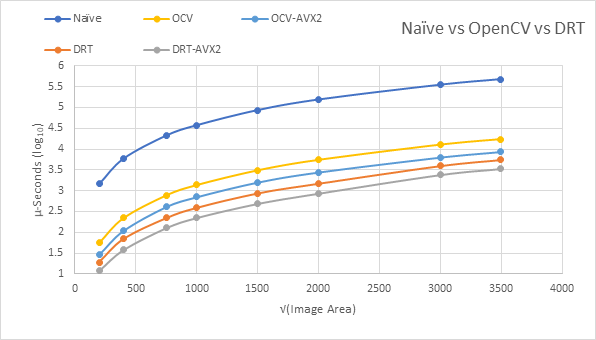}
\caption{Plot showing the execution time for the three algorithms as the image size increases. The execution time is in micro seconds (base 10).}
\label{fig:plot3}
\end{figure}

Figure \ref{fig:plot3} shows a plot comparing the time needed to compute the raw image moments of a given image size. The smallest image used was a 200x200 pixel image, with the largest at 4032x3024 pixels. The y-axis illustrates the execution time in micro-seconds (log base 10 for easier plotting). The x-axis is the square root of the number of pixels in the image (as this is a better method to illustrate the performance in terms of image size). We also include the OpenCV and DRT methods with SSE2/AVX optimisation (provided through the compiler settings). 

The DRT algorithm is faster, as is expected by it's complexity, the number of multiplications needed does not increase linearly with the number of pixels. We found the DRT algorithm was approximately 3 times faster than the OpenCV implementation, when both were used with the SSE2/AVX optimisations.

\section{Conclusion}

We have outlined an algorithm that utilises the Discrete Radon Transformation to significantly speed up the computation time for raw image moments of 2D grayscale images, outperforming the current algorithm that is used in computer vision libraries and in industry. This algorithm reduces the problem from computing 2D moments, to computing a set of 1D moments which are then linearly combined to give the correct results. We have derived the algorithm, showing that there is no loss of information and this method computes the exact raw image moments. It scales much better as the size of the input image increases, giving an improvement in execution time.

A possible generalisation of this algorithm is to 3 dimensional images (e.g. MRI scans) and whether is is possible to achieve $O(L+M+N)$ multiplication complexity for a pixel array of size $(L, M, N)$. Another avenue to consider is whether this can be extended to higher order moments as they have been shown to represent kurtosis \cite{prokop1992survey}.

\printbibliography

\appendix
\section{Image Projections}

Here we show an example of the projection arrays on an image of a brain scan. The following projection images display the values of each projection array at each point. The area bounded in white is the values from the projection arrays, forming a density plot of the projections.

\begin{figure}[ht]
\centering
\includegraphics[width=10cm, height=10cm]{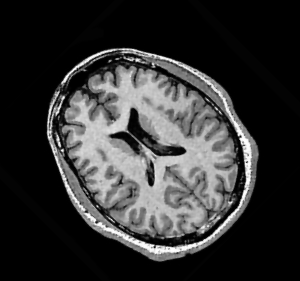}
\caption{Image of a brain.}
\label{fig:plot4}
\end{figure}

\begin{figure}
\centering
\frame{\includegraphics[width=.5\linewidth]{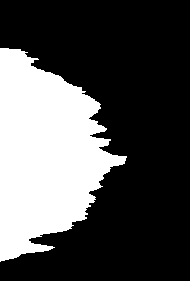}}
\caption{Horizontal projection of Figure 4}
\end{figure}

\begin{figure}
\centering
\frame{\includegraphics[width=.5\linewidth]{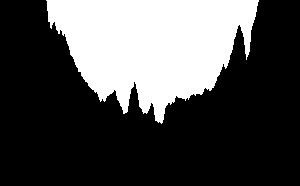}}
\caption{Vertical projection of Figure 4}
\end{figure}

\begin{figure}
  \centering
  \begin{minipage}[b]{0.4\textwidth}
    \frame{\includegraphics[width=\linewidth, height=15cm]{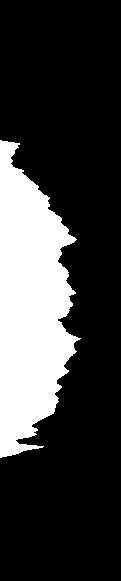}}
    \caption{Diagonal (45 degrees) projection of Figure 4}
  \end{minipage}
  \hfill
  \begin{minipage}[b]{0.4\textwidth}
    \frame{\includegraphics[width=\linewidth, height=15cm]{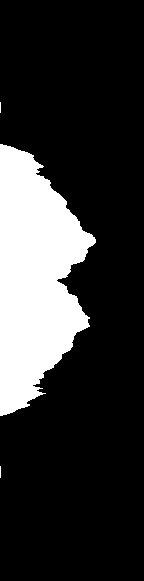}}
    \caption{Anti-Diagonal (135 degrees) projection of Figure 4}
  \end{minipage}
\end{figure}

\end{document}